\documentclass[letter,11pt]{article}
 
\usepackage{jheppub}
\usepackage{subcaption} 
\usepackage{graphicx}
\usepackage{comment}
\usepackage{dcolumn}
\usepackage{bm}
\usepackage{float}
\usepackage{hyperref}
\usepackage{dsfont}
\usepackage{slashed}
\usepackage{color}
\usepackage{amsmath}
 \usepackage{setspace}
\usepackage[section]{placeins}
\usepackage{braket}
\usepackage{upgreek}
\usepackage[bottom]{footmisc}
\usepackage[normalem]{ulem}
\usepackage{xcolor}
\usepackage{mathrsfs}
\usepackage[leftcaption]{sidecap}
\usepackage{subcaption}

\def\be{\begin{eqnarray*}}
\def\ee{\end{eqnarray*}}
\def\beq{\begin{eqnarray}}
\def\eeq{\end{eqnarray}}

\newcommand{\bea}{\begin{eqnarray}}
\newcommand{\eea}{\end{eqnarray}}

\usepackage{marginnote}

\title{Learning-Induced Dynamical Transition in Recurrent Neural Networks}

\author[a]{Varun Vaidya}

\affiliation[a]{\footnotesize Department of Physics, University of South Dakota, Vermillion 57069, USA}

\emailAdd{Varun.Vaidya@usd.edu}

\abstract{
Learning in recurrent neural networks can fundamentally reshape their underlying dynamics, transforming initially chaotic activity into stable task-dependent behavior. We develop a non-equilibrium dynamical mean-field theory(DMFT) to describe this transition during learning.  We show that a slow feedback-driven learning process generates an evolving effective feedback strength that drives the network through a transition from chaotic to stable dynamics defined by a bifurcation of the DMFT solution.  By deriving the two-time correlation function throughout learning, we identify a critical feedback strength and a corresponding learning rate dependent critical time separating these regimes. The transition arises from the progressive deformation of an effective dynamical landscape by the growing learned feedback structure.  Starting from the untrained state, the theory predicts the time evolution of the network output during training and shows quantitative agreement with numerical simulations. }

\begin{document}
\maketitle
\newpage

\newpage
\section{Introduction}

Recurrent neural networks provide a powerful framework for studying how complex dynamical systems can generate flexible and stable computational behavior. Beyond their role as models of artificial intelligence, recurrent networks have also become important theoretical models for understanding the principles governing high-dimensional biological neural circuits. A central question in this context is how learning modifies the intrinsic dynamics of recurrent networks, transforming initially irregular activity into structured dynamics capable of performing specific computational tasks.

Random recurrent networks exhibit a rich range of dynamical behaviors, including chaotic activity arising from the interaction between recurrent connectivity and nonlinear neuronal responses. Seminal work using dynamical mean-field theory \cite{sompolinsky1988chaos} established a framework for understanding these collective dynamical regimes and characterized transitions between chaotic and ordered states as a function of network parameters. Subsequent studies have shown that learning can substantially reshape these dynamics, either through changes in recurrent connectivity or through learned feedback pathways, leading to stable trajectories, attractor states, and task-dependent dynamical regimes \cite{Dauce1998,LajeBuonomano2013,SussilloAbbott2009,MastrogiuseppeOstojic2018,RivkindBarak2017}.

Recent theoretical studies have investigated how learned structure controls the dynamical state of trained recurrent networks. In particular, Clark et al. \cite{Clark2026} showed that sufficiently strong task-dependent recurrent restructuring can drive a transition between chaotic and stable regimes. Such approaches provide a phase-space description of learned networks, identifying the conditions under which stable computation becomes possible.

However, an important question remains: how does learning itself dynamically drive a network from one dynamical regime to another? During training, the network is not a stationary system. Its synaptic parameters evolve continuously, modifying the effective recurrent dynamics and, consequently, the statistical properties of neural activity. Understanding this process therefore requires following the full non-equilibrium learning trajectory, rather than characterizing only the final trained state.
In this work, we develop a dynamical mean-field theory for recurrent neural networks undergoing slow, feedback-driven learning. Rather than introducing an externally controlled parameter to tune between dynamical regimes, we show that learning itself generates an evolving effective control parameter through the gradual formation of structured feedback. This evolving feedback dynamically deforms the effective landscape governing network fluctuations, driving the system through a transition from chaotic to stable dynamics.

A central object in our analysis is the full two-time correlation function between neuronal outputs, which captures the non-stationary nature of the learning process. Building on the two-time correlation and response formalism of dynamical mean-field theory \cite{sompolinsky1988chaos}, we use this description to track the emergence of long-time correlations and identify a critical feedback and a corresponding learning rate dependent critical time at which the dynamical regime changes. We show that this transition corresponds to a qualitative change in the effective dynamical landscape, where the evolution of learned feedback drives the system across a critical point. Quantitatively this corresponds to a bifurcation of the DMFT solution. We show that during learning, the two time correlation function exhibits a separation between rapid chaotic relaxation and slow evolution of the correlation plateau, allowing the learning trajectory to be described as a slow deformation of the underlying dynamical state.

Beyond characterizing the transition itself, our theory predicts the evolution of the network output throughout learning. The predictions obtained from the dynamical mean-field equations show quantitative agreement with numerical simulations, demonstrating that the theory captures both the transient learning dynamics and the emergence of the final stable computational state.

Our results provide a framework for understanding learning as a dynamical process that can reorganize the dynamical regime of a recurrent system. Rather than viewing stable computation as a property imposed by a fixed learned structure, we show how stability can emerge dynamically through slow plastic adaptation, providing a mechanism by which recurrent networks transition from internally generated chaotic activity to reliable task-dependent dynamics.

The paper is organized as follows.  In Section \ref{sec:Model}, we introduce the slow feedback-driven learning model for our recurrent neural  network.  We develop the DMFT analysis for this system in Sections \ref{sec:DMFT} and \ref{sec:Pot}.  The time evolution of the system during training is derived in Section \ref{sec:Tevol} followed by a comparison with simulation in Section \ref{sec:Num}.  The conclusions along with an outlook are presented in Section \ref{sec:Conclusion}.
\section{The learning model for RNN}
\label{sec:Model}

We consider a recurrent neural network consisting of (N) interacting units whose dynamics are governed by
\begin{equation}
\tau_c \dot{x}_i(t)=-x_i(t)+g^2\sum_{j=1}^{N}J_{ij}\phi(x_j(t))+\sum_{\mu}W^{fb}_{i}y(t),
\end{equation}

where $x_i(t)$ denotes the activity of neuron (i),  $\phi(x)$ is a nonlinear activation function, and $J_{ij}$ represents the recurrent connectivity.  $\tau_c$ is the characteristic time scale for the reservoir dynamics. Throughout this paper, we use the standard activation function $\phi(x) = \tanh{x}$. 
The recurrent couplings are drawn from a Gaussian distribution with zero mean and a variance $1/\sqrt{N}$, and we analyze the statistics of this ensemble.

The network produces task-relevant outputs through a low-dimensional readout,
\begin{equation}
y(t)=\sum_i w^{out}_{i}(t)\phi(x_i(t)).
\end{equation}

These outputs are fed back into the recurrent network through feedback projections $W^{fb}_{i}$ which are also drawn from a Gaussian distribution with mean zero and variance $\sigma_{fb}/\sqrt{N}$. Feedback from the readout to the recurrent reservoir is a standard architecture in closed-loop reservoir computing and FORCE-based recurrent networks \cite{Jaeger2001,SussilloAbbott2009, Koryakin2012}, where it enables recurrent networks to autonomously generate learned trajectories and complex temporal dynamics. Here we adopt this architecture as a biologically motivated mechanism through which task-related information can progressively reshape recurrent dynamics during learning. From a biological perspective, the feedback can be interpreted more generally as a top-down signal that modulates ongoing neural activity and conveys task-relevant information \cite{Murray2019,Miconi2017,AsabukiClopath2025}.

The central feature of the model is that the readout weights evolve slowly according to a learning rule. We consider a feedback-driven learning process in which the readout is modified according to the difference between the desired target signal ($y^{*}(t)$) and the generated output,
\begin{equation}
e(t)=y^{*}(t)-y(t),
\label{eq:error}
\end{equation}

with synaptic adaptation governed by
\begin{equation}
\dot{w}^{out}_{i}(t)=\frac{1}{N}\alpha \ e(t)\phi(x_i(t)).
\label{eq:wout}
\end{equation}
Here ($\alpha \ll 1/\tau_c $) sets the slow learning timescale relative to the intrinsic neural dynamics. This separation of timescales allows the recurrent activity to relax rapidly for a given value of the slowly evolving synaptic parameters, while learning gradually reshapes the effective network dynamics.

The purpose of this work is not only to determine the final learned output, but to understand how the dynamical state of the network evolves during learning. Initially, the random recurrent network exhibits chaotic fluctuations. As the feedback pathway becomes increasingly structured through learning, it modifies the effective dynamics of the recurrent system. We therefore study the non-equilibrium learning trajectory generated by the coupled evolution of neural activity and synaptic parameters.

The main object characterizing this evolution is the two-time correlation function,
\begin{equation}
C(t,s)=\frac{1}{N}\sum_i\phi(x_i(t))\phi(x_i(s)),
\end{equation}
which captures the non-stationary dynamics of the learning process. Unlike equilibrium or fully trained analyses, where correlations depend only on the time difference (t-s), learning generates a system whose statistical properties evolve with time. This two-time structure allows us to follow the emergence of long-time correlations starting from a chaotic state and the transition between dynamical regimes during training.
\section{Disorder averaging and DMFT analysis}
\label{sec:DMFT}

The MSRJD path integral framework  \cite{MartinSiggiaRose1973,Janssen1976,DeDominicis1976} applied  to random neural networks \cite{CrisantiSompolinsky2018}  introduces response fields $\tilde{x}_i(t)$ to encode the dynamics through a generating functional  written as a path integral:
\begin{equation}
Z = \int \prod_i \mathcal{D}x_i \mathcal{D}\tilde{x}_i \; \exp \Bigg\{
i \sum_i \int dt \, \tilde{x}_i(t) \Big[ \tau \dot{x}_i(t) + x_i(t) - \sum_j \left(g^2 J_{ij}+W_i^{fb}w_j^{out}\right) \phi(x_j(t))  \Big] 
\Bigg\}.
\end{equation}
The response fields $\tilde x_i(t)$  enforce the microscopic equations of motion within the generating functional. 
We can perform an ensemble average over the network parameters $J_{ij}, W_i^{fb}$, using the distributions 
\begin{eqnarray}
    P(J_{ij}) = \frac{1}{\sqrt{2\pi N}}\exp{\{-\frac{N}{2g^2}J_{ij}^2\} }, \ \ \ \  P(W_i^{fb}) = \frac{\sigma_{fb}}{\sqrt{2\pi N}}\exp{\Big\{-\frac{N}{2\sigma_{fb}^2}(W^{fb}_{i})^2\Big\} }
\end{eqnarray}
Following details presented in Appendix \ref{app:DMFT}, this yields an effective  one neuron equation
\begin{eqnarray}
    \tau_c \dot{x}_i(t)+x_i(t) = \eta_i(t)
    \label{eq:oneneuron}
\end{eqnarray}

where $\eta_i(t)$ is the effective noise term that obeys Gaussian statistics encoded in the effective action 
\begin{eqnarray}
    S_{\text{eff}} = -\frac{1}{2}\int dt ds \sum_i \eta_i(t) \bar C^{-1}(t,s) \eta_i(s)
    \label{eq:Seff}
\end{eqnarray}
where 
\begin{eqnarray}
    \bar C(t,s) &=& \frac{g^2}{N}\sum_i \langle \phi(x_i(t)) \phi(x_i(s))\rangle +\frac{\sigma_{fb}^2}{N}\langle y(t)y(s)\rangle \nonumber\\
    &\equiv & g^2 C(t,s) + \frac{\sigma_{fb}^2}{N}\langle y(t)y(s)\rangle
\end{eqnarray}
is a correlator which is determined self consistently using this action.We can further simplify this equation in the large N limit by noting that 
\begin{eqnarray}
\langle y(t) y(s) \rangle = \sum_{i,j} \langle w_{i}^{out}(t)w_j^{out}(s) \phi(x_i(t))\phi(x_j(s))\rangle 
\end{eqnarray}
$w_i^{out}$ itself evolves through Eq.~\ref{eq:wout}, which is proportional to the activity of the $i^{th}$ neuron. 
Since cross correlations between distinct neurons are suppressed in the large N limit, the dominant contribution for evolution is $ \langle y(t) \rangle \langle y(s) \rangle $. The object of interest is then $\langle y(t) \rangle $.
Given Eqns.~\ref{eq:error}, \ref{eq:wout}, we can write
\begin{eqnarray}
    w_i^{out}(t) = \frac{\alpha}{N}\int_0^t ds (y^*(s) -y(s) )\phi(x_i(s)) 
\end{eqnarray}
so that 
\begin{eqnarray}
    \langle y(t) \rangle = \frac{\alpha}{N}\int_0^t ds \sum_i \langle (y^*(s) -y(s) )\phi(x_i(s)) \phi(x_i(t))\rangle 
\end{eqnarray}
which again, in the large N limit reduces to 
\begin{eqnarray}
    \langle y(t) \rangle = \alpha\int_0^t ds (y^*(s)-\langle y(s)\rangle )C(t,s)
    \label{eq:out}
\end{eqnarray}
Clearly, to obtain a closed system of equations, we need an evolution equation for $C(t,s)$ which can be obtained given its definition and the fact that $\eta_i(t)$ evolves with the action Eq.~\ref{eq:Seff}. 
We set $\tau_c =1$ henceforth so all time scales are measured in units of $\tau_c$. Formally Eq.~\ref{eq:oneneuron} can be solved to write 
\begin{eqnarray}
    x_i(t) = \int^t dt' \exp{\{t'-t\}} \eta_i(t) 
\end{eqnarray}
Using this result, we can define the autocorrelation function $\Delta(t,s)$ of the neuronal activity
\begin{eqnarray}
  \Delta(t,s)\equiv  \langle x_i(t) x_i(s) \rangle = \int^t dt' \exp{\{t'-t\}} \int^s ds' \exp{\{s'-s\}} \langle \eta_i(t) \eta_i(s) \rangle
\end{eqnarray}
which given Eq.~\ref{eq:Seff} obeys the two-time evolution equation 
\begin{eqnarray}
    (1+\partial_s)(1+\partial_t) \Delta(t,s) = g^2C(t,s) +\frac{\sigma_{\text{fb}}^2}{N}\langle y(t) \rangle \langle y(s) \rangle 
    \label{eq:autoC}
\end{eqnarray}
Since $x_i(t)$ evolves with Gaussian statistics,  $C(t,s)$ can be expressed as a function of the autocorrelation function $\Delta(t,s)$  \cite{sompolinsky1988chaos}
\begin{eqnarray}
    C(t,s) =  \int Dz Dy Dx \phi(\sqrt{\Delta(t,t) -\Delta(t,s)}x+z \sqrt{\Delta(t,s)})\phi(\sqrt{\Delta(s,s)-\Delta(t,s)}y+z \sqrt{\Delta(t,s)}) \nonumber \\
    \label{eq:Cts}
\end{eqnarray}
where $\int Dz \equiv \int dz/\sqrt{2\pi}\exp{\{-z^2/2\}}$ is the integral over a normal distribution. Eqns.~\ref{eq:out}, \ref{eq:autoC} and \ref{eq:Cts} form a closed system of equations that can, in principle be solved numerically to obtain the the complete evolution of the system in time. However, we see that they are highly non linear and due to the triple integral in the definition for $C(t,s)$, a brute force solution would be expensive. Likewise, a direct numerical approach also would offer little intuition about the underlying physics.
Therefore, in the next section we will explore the underlying physics dictated by these equations by exploiting the separation in time scales, which will give us some insight at the cost of quantitative precision. To gain insight and simplify the analysis, we will the set the target to a constant value $y^*(t) = A \sim O(\sqrt{N})$.The scaling is chosen so that the feedback term, when the network is fully trained, is of the same order as the internal chaotic dynamics and is always relevant. We can consider two other regimes:1.  $ A \gg \sqrt{N}$ in which case we can expect the system to quickly align with the output since it will overwhelm any internal fluctuations. The case where $A \ll \sqrt{N}$ is also interesting and we will comment on this in Section \ref{sec:Conclusion}.  
\section{Learning as a dynamical deformation of an effective potential}
\label{sec:Pot}
One of the central quantities in the set of equations derived in the previous section is the equal time auto-correlation function $\Delta(t,t)$.  Given the separation of time scales $1/\alpha \gg \tau_c $, if we restrict $t-s \ll 1/\alpha$, this implies looking at $\Delta(t,s)$ matrix near the diagonal. For this part of the matrix, we can adopt an SCS-type analysis defining a central time $T= (t+s)/2$ and $\tau = t-s$.  Since evolution over central time is slow, we can ignore any derivatives with respect to $T$  which scale as $\alpha \ll 1$. Likewise the output evolves over a time scale $1/\alpha \gg 1$, so that it is effectively constant across $\tau$. With these approximations, Eq.~\ref{eq:autoC} simplifies to
\begin{eqnarray}
    (1-\partial^2_{\tau}) \Delta(T,\tau) = g^2 \int Dz \Big[ Dx \phi(\sqrt{\Delta(T) -\Delta(T,\tau)}x+z \sqrt{\Delta(T,\tau)})\Big]^2+ \frac{1}{N}\langle y(T) \rangle^2
\end{eqnarray}
We can define a normalized output $\hat y(T) = 1/\sqrt{N} \langle y(T) \rangle$, choosing for simplicity $\sigma_{fb}= 1$. Since we are assuming a constant target $y^*(t) = A$, the output of the network $\hat y(T)$, whenever learning is successful, is a monotonic function of time. Hence, we can consider $\Delta(T, \tau)$ as a function $\Delta(\hat y, \tau)$,
\begin{eqnarray}
    (1-\partial^2_{\tau}) \Delta(\hat y,\tau) = g^2 \int Dz \Big[ Dx \phi(\sqrt{\Delta(\hat y,0) -\Delta(\hat y,\tau)}x+z \sqrt{\Delta(\hat y,\tau)})\Big]^2+ \hat y^2 
    \label{eq:fastDelta}
\end{eqnarray}
Therefore, at least near the diagonal, the problem reduces to solving the SCS equation with a deformation. Following \cite{sompolinsky1988chaos}, we can adopt the picture of a particle moving in an effective potential which slowly evolves with time but can be considered as stationary over the time scale of the reservoir. In this case, $\hat y$ evolves monotonically from 0 to $A/\sqrt{N}$. We note again that this picture is only valid when $t-s \ll 1/\alpha$ which is a narrow band around the diagonal of the matrix $\Delta(t,s)$. We call this solution $\Delta_{\text{fast}}(\hat y,\tau)$. Defining $u = \Delta_{\text{fast}}(\hat y, \tau)$, the equation reduces to 
\begin{eqnarray}
    \ddot{u} = -\frac{dV(u,\hat y)}{du}  \ \ \text{where} \ V(u, \hat y)= -\frac{1}{2}u^2+g^2 \int_0^u dv \int Dz \Big[ Dx \phi(\sqrt{\Delta(\hat y,0) -u}x+z \sqrt{u})\Big]^2+ \hat y^2 u \nonumber\\
\end{eqnarray}

\begin{figure}
\centering
\includegraphics[width=\linewidth]{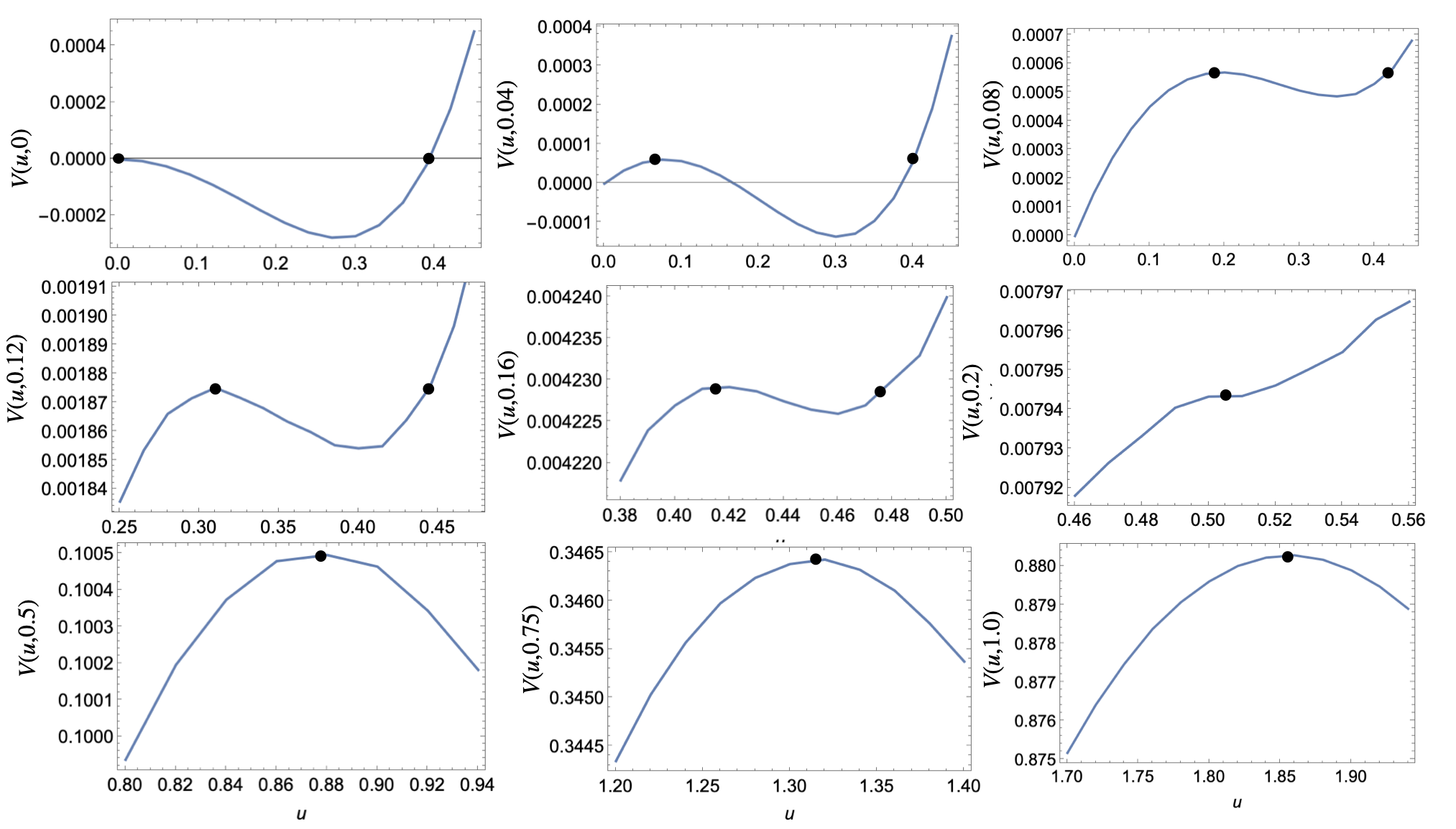}
\caption{The deformation of the effective one dimensional potential as a function of the feedback parameter $\hat y$ for g=1.3. Starting from the SCS picture in (a), increasing feedback with time deforms the potential to a narrower and shallower well with the simultaneous emergence of a non- zero plateau. At $\hat y = 0.2$,  subplot (f), we see the collapse  of the well signaling a qualitative change. Beyond this, the solution is frozen at the maxima of the potential and fast dynamics disappear.}
\label{fig:potpanel}
\end{figure}

We know that for the original SCS equation ($\hat y=0$), the stable solution is a monotonic decay from the g-dependent initial value $\Delta(0,0)$ to zero for $\tau \gg \tau_c $. Within the mechanical analogy of the dynamical mean-field equation, this corresponds to the motion of a particle through the effective potential, evolving from $u= \Delta(0,0)$ to the stationary point at u=0, as illustrated in Fig.~\ref{fig:potpanel}(a).

As  $\hat y$ increases with time, the potential becomes progressively shallower while the width of the well decreases. At the same time, the long-time stationary point $\Delta(\hat y, \infty)$ moves continuously to a nonzero u, reflecting the emergence of a non-zero plateau in the correlation function. Most importantly, at a critical value  of $\hat y$, the local minimum and adjacent maximum merge, causing the potential well to collapse and the system moves through the stability landscape. Beyond this critical time $t_{\text{cr}}$, the fast-time fluctuations can no longer be sustained, and the system enters the stable regime. This sequence is illustrated in Fig.~\ref{fig:potpanel} for a representative value of g=1.3. 
Similar transitions between the chaotic and stable regimes have recently been obtained by varying an externally imposed feedback parameter $\gamma$ and studying the asymptotic state of the network \cite{Clark2026}. In contrast, here no external control parameter is introduced. Instead, the learning dynamics itself continuously reshapes the effective potential, driving the network across the bifurcation at a finite critical feedback and hence a critical learning time.

This interpretation immediately predicts that the disappearance of the potential well should coincide with the collapse of the fast-time component of the two-time correlation function, a prediction that is confirmed by the numerical solution presented below.

The condition that the minima and maxima merge at the critical feedback means that $\Delta(\hat y,0) = \Delta(\hat y, \infty) =u_c$ obeys
\begin{eqnarray}
\frac{d^2}{du^2}V(u, \hat y)= 0   \implies 1= g^2 \frac{d}{du} \int Dz  \Big[\int Dx \phi(\sqrt{u_c-u}x+z\sqrt{u})\Big]^2 \Big|_{u=u_c}
\end{eqnarray}

Expanding about $u= u_c$, and simplifying we arrive at 
\begin{eqnarray}
    1=  g^2 \int Dz \phi'^2(\sqrt{u_c}z)
    \label{eq:stab}
\end{eqnarray}
Interestingly, this is identical to the marginal stability condition obtained in the SCS analysis\cite{sompolinsky1988chaos} of random recurrent networks. In the original SCS setting, this condition determines the boundary between chaotic and fixed-point dynamics as the recurrent coupling strength g is varied. Here, however, g is fixed and the transition is driven dynamically by learning. The evolving feedback changes the stationary solution $u_c(\hat y)$ through the self-consistency condition
\begin{eqnarray}
\frac{d}{du}V(u, \hat y) = 0   \implies u_c= g^2 \int Dz \phi^2(\sqrt{u_c}) + \hat y_c^2
\label{eq:uc}
\end{eqnarray}
causing the system to move through the stability landscape until it reaches the marginal point $u_c$.
This occurs when the output of the network reaches a specific value $\hat y$. As shown in Fig.\ref{fig:potpanel} (f), for g=1.3, this number is $\hat y= 0.2$ and in general will increase with increasing value of g.  We notes that this number is independent of the learning rate $\alpha$. On the other hand, the specific time at which this happens during the learning trajectory defines a critical time $t_{\text{cr}}$
which will be a function of g and $\alpha$. This requires us to solve explicitly for the time dependence of $\hat y(t)$, which we do in Section \ref{sec:Tevol}.

Thus, the freezing transition does not arise from a change in the local stability criterion itself, but from the learning-induced evolution of the correlation structure that brings the system to the SCS marginal state. Solving Eq.~\ref{eq:fastDelta} numerically allows us to compute $\Delta(\hat y, \tau)$ and consequently $C(\hat y, \tau)$ for $\tau \ll 1/\alpha$. The numerical results will be presented later in Section \ref{sec:Num} after we finish the analysis for the far off diagonal elements and the time evolution of $\hat y$. 

Now we consider the off-diagonal elements of $\Delta(t,s)$ in the region $t-s \geq 1/\alpha \gg \tau_c $. The fast correlations have decayed away so all that remains is a slow evolution of the plateau. Hence all the time derivatives in Eq.~\ref{eq:autoC} scale as $\alpha \ll 1$ and can be ignored. In this case the two-time evolution equation reduces to a self consistent equation
\begin{eqnarray}
    \Delta(p,q) &=& g^2  \int Dz Dy Dx \phi(\sqrt{\Delta(p,p) -\Delta(p,q)}x+z \sqrt{\Delta(p,q)})\phi(\sqrt{\Delta(q,q)-\Delta(p,q)}y+z \sqrt{\Delta(p,q)})\nonumber\\
    &+& pq
    \label{eq:slowD}
\end{eqnarray}
where $p \equiv \hat y(t) $, $q \equiv \hat y(s)$. We have already solved for the diagonal value $\Delta(p,p), \Delta(q,q)$ which then allows us to solve for the far off diagonal correlation function which we call $\Delta_{\text{slow}}(p,q)$. The corresponding two time correlation function  $C_{\text{slow}}(p,q)$ can then be computed as well.

Given the separation of time scales, we now approximate the full $\Delta(t,s)$ matrix as follows. Starting from the diagonal value, we have a fast decay to the plateau dictated by the potential as shown in Fig.~\ref{fig:potpanel}. This gives us $\Delta_{\text{fast}}(t,\tau)$ over a time scale $t-s \ll 1/\alpha$. We then match this to a slowly evolving plateau $\Delta_{\text{slow}}(t,s)$, over time scales $t-s \geq 1/\alpha$ dictated by Eq.~\ref{eq:slowD}. 
The matrix $C(t,s)$ is approximated in exactly the same manner allowing us to write
\begin{eqnarray}
 C(t,s) \approx \left( C_{\text{fast}}(t,\tau)  -C_{\text{fast}}(t,\tau= \infty) \right) + C_{\text{slow}}(t,s)
 \label{eq:Cmatrix}
\end{eqnarray}

So far, we have this matrix as a function of $\tau$ and $\hat y$ for short time scales around the diagonal and as function of $\hat y(t), \hat y(s)$ for longer times. We still need to solve for $\hat y$ as a function of time to obtain the full time dependence which we turn to in the next section.

\section{Time evolution of the system}
\label{sec:Tevol}

 In this section, we explicitly solve for the learning trajectory as a function of time. There are four time scales in the problem, two of  which are the initial parameters of the system namely $\tau_c$ and $1/\alpha$ which govern the fast dynamics and the slow feedback learning respectively. We also have two emergent scales, $t_{\text{decay}}$, the time scale over which correlators undergo a fast decay to a plateau. Finally we have $t_{\text{cr}}$ when the system transitions from chaotic to stable dynamics. This naturally allows us to divide the time evolution into three regimes as follows.
\subsection{Evolution at early time}

\begin{figure}
\centering
\includegraphics[width=\linewidth]{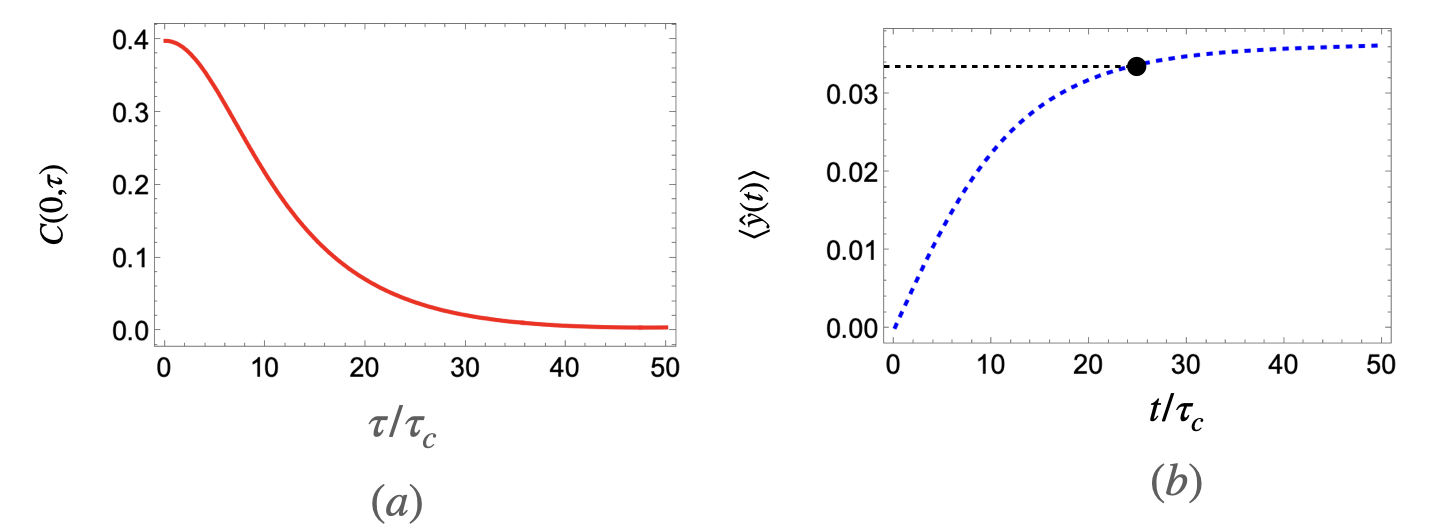}
\caption{Evolution at early time $t \ll 1/\alpha$. (a) shows the fast decay of the SCS correlator $C(0,\tau)$. (b) shows the time evolution of the network output at early times dictated by this fast decay.}
\label{fig:early}
\end{figure}

At early times, $t  \leq t_{\text{decay}} \ll 1/\alpha $ when the output is small, the feedback is not strong enough to overcome the chaotic fluctuations completely. Hence the dynamics is sensitive to the fast fluctuations over the time scale $\tau_c$. The evolution of the output is governed by Eq.~\ref{eq:out}. Since the integral over s is capped by t, in this regime, the off diagonal correlation matrix C(t,s) is relevant only over the interval $t-s  \ll 1/\alpha $. Similarly the diagonal value $\Delta(t,t) \approx \Delta(0,0)$ is almost a constant over this small time interval since no appreciable learning has happened yet. Therefore the system is essentially in an SCS chaotic state. In that case we can write
\begin{eqnarray}
\langle y(t) \rangle  \approx \alpha A \int_0^t ds C_{\text{fast}}(t=0, t-s)
\end{eqnarray}
where we have also assumed that $\langle y(t) \rangle $ is small and so the error is approximately given by the target A. 
\begin{eqnarray}
    &&\langle y(t) \rangle = A \alpha \chi(t) \ \ \text{where}  \ \ \chi(t) = \int_0^t ds C_{\text{fast}}(t=0, t-s)
\end{eqnarray}
To proceed further we make explicit choices $A = \sqrt{N}$ and $\alpha = 0.007$. This ensures that the feedback to the reservoir when it is fully trained is O(1) and $\alpha$ is sufficiently small for our separation of time scales to hold. Our solution for the early time evolution is then $\hat y(t) =\alpha \chi(t)$ and is shown in Fig.~\ref{fig:early}(b). 
We see in Fig.~\ref{fig:early}(a) that the SCS fast decay happens over a times scale of $t_{\text{decay}} \sim $ 20-30 $\tau_c$ . This is much smaller than $1/\alpha  \approx 140 \tau_c$. We want to use this approximate solution for $\hat y$ upto a time where the SCS potential has not deformed appreciably. So we choose $\hat y(25) =.033$ as the boundary for this early time evolution where the fast correlator has decayed to the plateau. We see that over the time interval $\{0,25\} \tau_c$, when $\hat y$ rises from 0 to $0.033$, the potential is still very close to the SCS potential as shown in Fig.\ref{fig:potpanel}. $\Delta(t, \tau=0) \approx 0.4$, while $\Delta(t,\infty)$ rises by a very small value to $0.05$. 
\subsection{Evolution at intermediate time}

We now consider the intermediate-time regime which corresponds to the interval $t_{\text{cr}} \geq  t \geq t_{\text{decay}}$. In this regime, both fast and slow dynamics are important. As explained in the last section, given the separation of scales, we can approximate the correlation matrix as a matched solution of the fast and slow matrix Eq.~\ref{eq:Cmatrix}. This enables us to write the output using Eq.~\ref{eq:out},
\begin{eqnarray}
    \langle y(t) \rangle = \alpha\int_0^{t} ds \left(A- y(s) \right)C_{\text{slow}}(t,s)+ \left(A- y(t) \right) \alpha\int_{0}^t ds (C_{\text{fast}}-C_{\text{fast}}(t,\tau= \infty))
\end{eqnarray}
Since the second term, governed by fast dynamics, only has support over time scale $ t-s \ll 1/\alpha$, the output essentially remains constant at $y(t)$ and hence can be pulled out of the integral. This also allows us to set the upper limit of integration for this second term to $\infty$ since in this regime we are only looking at $ t \geq t_{\text{decay}}$. 
Since $\langle y(t) \rangle $ is a monotonic function of time, we can make a change of variables to $p =\langle y(t) \rangle$ in the first term which allows us to write 
\begin{eqnarray}
    p = \alpha \int_0^p dv (A- q) f'(q) C_{\text{slow}}(p,q) + \alpha (A-p) K(p)
\end{eqnarray}
where $f(p) \equiv y^{-1}(p)=t$  and the kernel $K(p)$ is known completely from the fast dynamics 
\begin{eqnarray}
    K(p) = \int_0^{\infty} d \tau \left( C_{\text{fast}}(p,\tau)- C_{\text{fast}}(p,\tau= \infty)\right)
\end{eqnarray}
We again note that since the integrand in the definition of K(p) goes to zero over a time scale $t_{\text{decay}}$, we can safely take the limit of integration to $\infty$ when looking at evolution at time t larger than the typical fast dynamics  decay time. We have then reduced the problem to solving for the function $f(p) = y^{-1}(p) = t$, which when inverted gives us the solution for the learning trajectory. This can be done by discretizing the integral and implementing a finite difference for the derivatives 
\begin{eqnarray}
    p_n = \alpha \sum_{i=1}^n  (A- p_i)(f_{i+1}-f_i)C_{\text{slow}}(p_n,p_i)+ \alpha(A- p_n)K(p_n)
\end{eqnarray}
Rearranging, we can solve for $f_{n+1}$ given all values till $p_n$ and $f_n$ through the equation 
\begin{eqnarray}
    f_{n+1}= f_{n}+ \frac{u_n -\alpha (A-p_n)K(p_n)-\alpha \sum_{i=1}^{n-1} (A-p_i) (f_{i+1}-f_i)C_{\text{slow}}(p_n,p_i)}{\alpha (A-p_n)C_{\text{slow}}(p_n,p_n)}
    \label{eq:intimes}
\end{eqnarray}
Since we know $p= \langle y(t) \rangle $ at early times, upto the decay of fast dynamics ($\sim t= 25 \tau_c$ for $g=1.3$), we already know the function $f(p) = t$ till that time. This is the initial condition for Eq.~\ref{eq:intimes} which we can then march forward to solve for all subsequent values of $f_i$. 

\subsection{Evolution at late time}
As the output of the network increases, chaotic fluctuations are increasingly suppressed, until they completely disappear beyond $t =t_{\text{cr}}$. Hence, beyond this time, the kernel $K(u)$ goes to zero and we only have contribution from slow evolution. So we can continue using Eq.~\ref{eq:intimes} setting $K(u)$ to 0 beyond $u= u_c$(Eqns. \ref{eq:stab} and \ref{eq:uc}) ,  which is 0.2 for g=1.3.
\begin{eqnarray}
    f_{n+1}= f_{n}+ \frac{u_n-\alpha \sum_{i=1}^{n-1} (A-u_i) (f_{i+1}-f_i)C_{\text{slow}}(u_n,u_i)}{\alpha (A-u_n)C_{\text{slow}}(u_n,u_n)}
    \label{eq:itimes}
\end{eqnarray}

This now allows to solve for the function f throughout the learning process, allowing us to access the temporal evolution of the network output as well as the two-time correlation function.
\section{Simulation and comparison with DMFT}
\label{sec:Num}
\begin{figure}
\centering
\includegraphics[width=\linewidth]{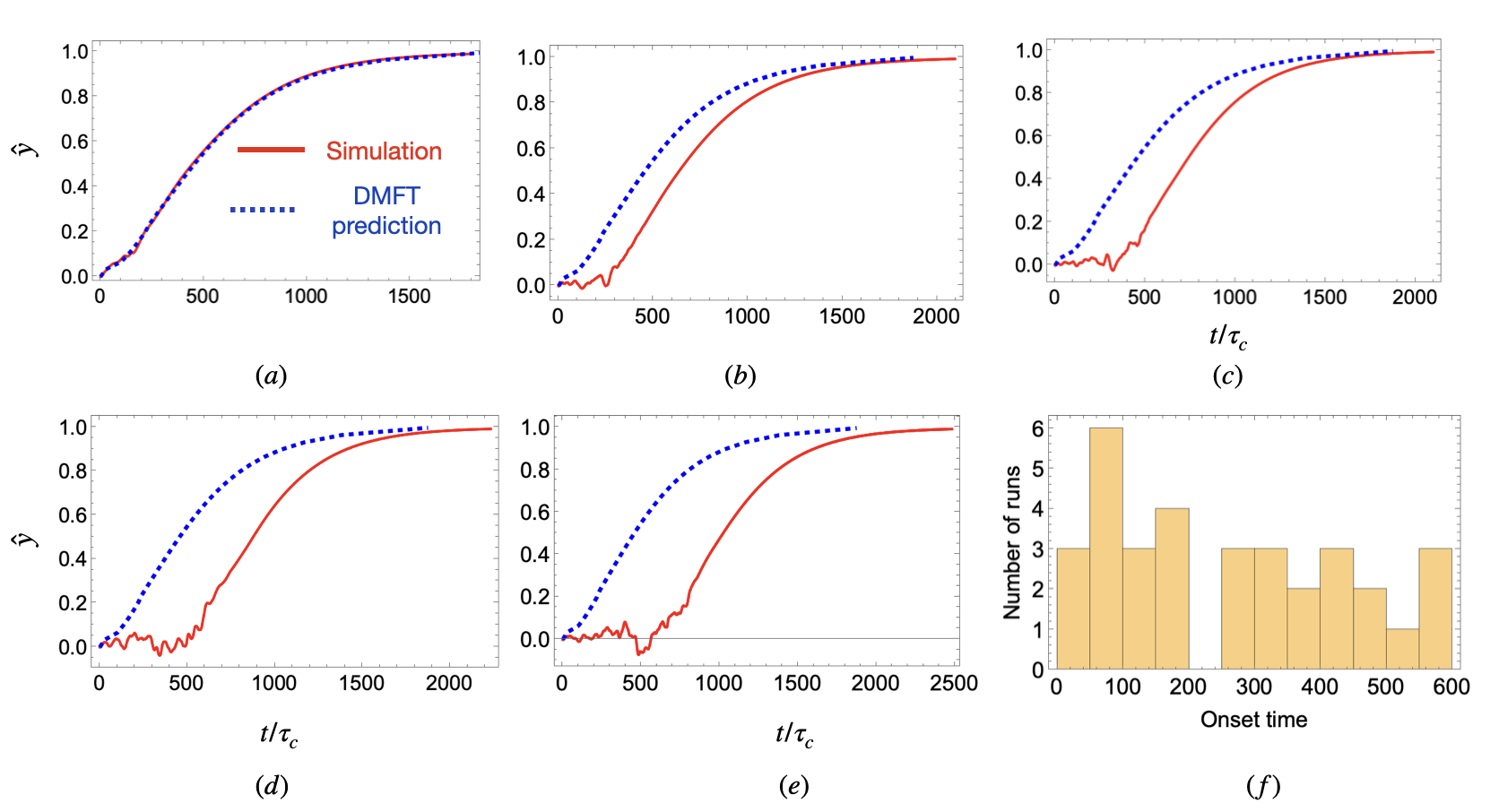}
\caption{Network output as a function of time for five representative realizations with different onset times. The dashed curve shows the DMFT prediction. The histogram in (f) shows the distribution of onset times over 33 realizations.}
\label{fig:ztrajpanel}
\end{figure}

Now we present the numerical results comparing simulation with the DMFT prediction. To validate our theory, we choose a representative value of g=1.3 simulated for a network of N= 5000  neurons and a constant target $A= \sqrt{N}$. The network was initially prepared in an SCS state by evolving it  without any feedback.  
Fig.~\ref{fig:ztrajpanel} shows 5 representative runs along with the DMFT prediction after the feedback and learning are implemented. 
We note that at finite N, for each run there is a variable time lag before macroscopic learning commences. We also show the histogram for the onset time of learning over 33 runs  in Fig.\ref{fig:ztrajpanel}. The DMFT gives a deterministic trajectory and so does not capture run to run variability in the onset time of learning. We can clearly identify the critical time $t_{\text{cr}}$ from the output trajectory as the time when the fluctuations completely disappear and the output rises smoothly. This happens at $\hat y= 0.2$ for g=1.3 as predicted by the DMFT analysis.
\begin{figure}
\centering
\includegraphics[width=\linewidth]{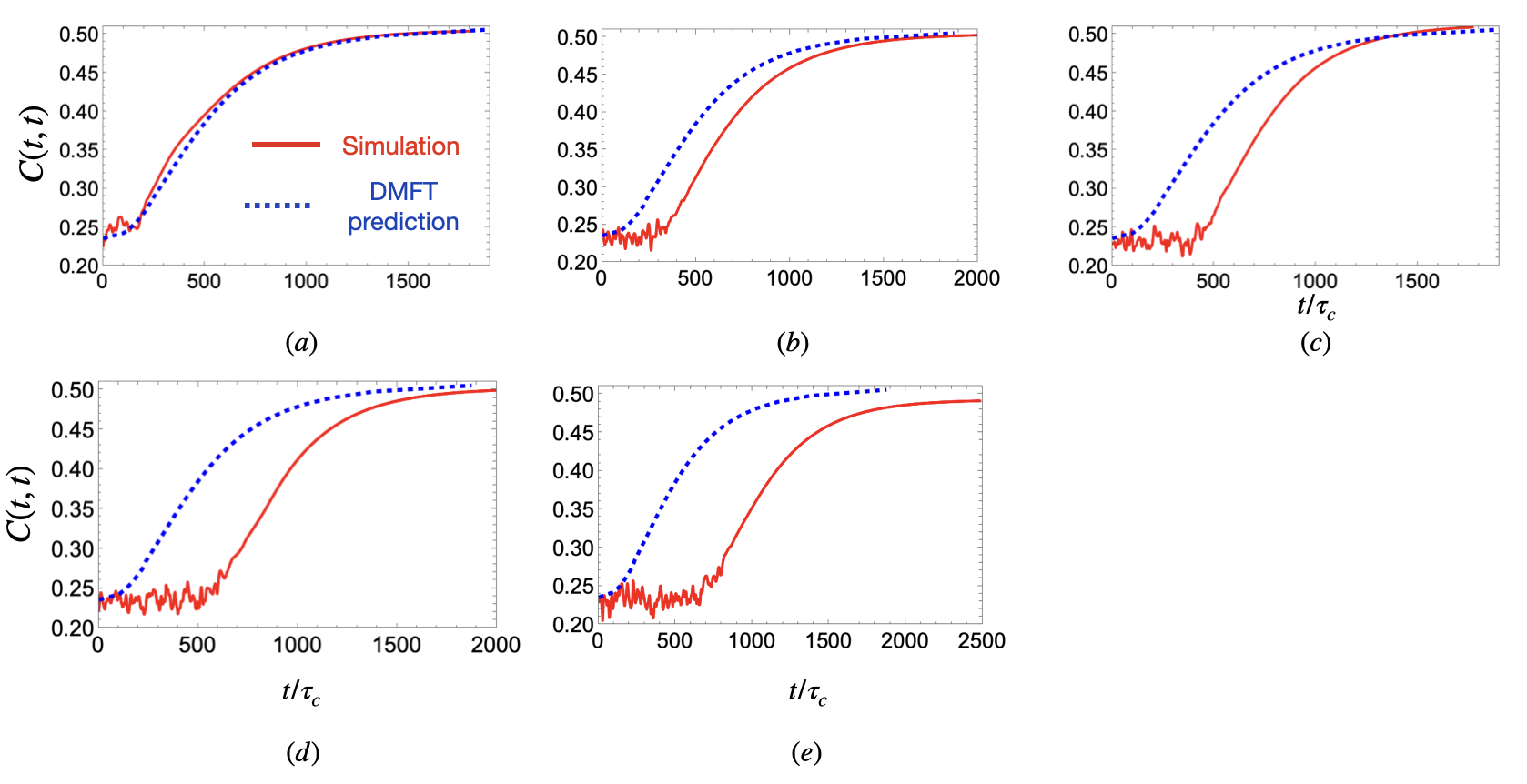}
\caption{Diagonal correlation C(t,t) for five representative realizations. The dashed curve shows the DMFT prediction.}
\label{fig:Cpanel}
\end{figure}
Fig.~\ref{fig:Cpanel} shows the diagonal entries of the correlation matrix $C(t,t)$ for the same representative runs. We see the the same time lag after which the correlation matrix starts rising above the SCS value of 0.23. We also overlay the DMFT prediction for comparison. 

\begin{figure}
\centering
\includegraphics[width=\linewidth]{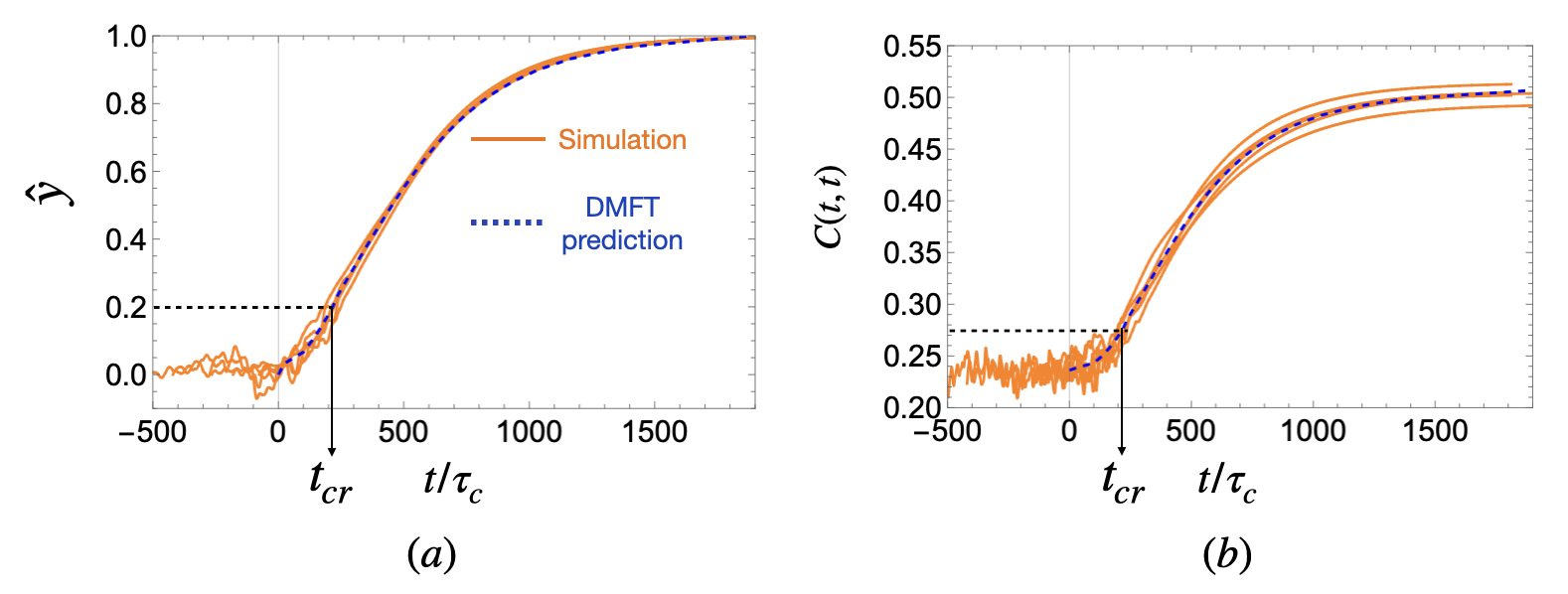}
\caption{(a) Network output as a function of time for representative runs shifted by their onset times. Also shown is the critical time $t_{\text{cr}}$ as predicted by the DMFT (b) The Diagonal Correlation matrix $C(t,t)$ as a function of time for the same representative runs shifted by their onset times. }
\label{fig:shifted}
\end{figure}
\begin{figure}
\centering
\includegraphics[width=\linewidth]{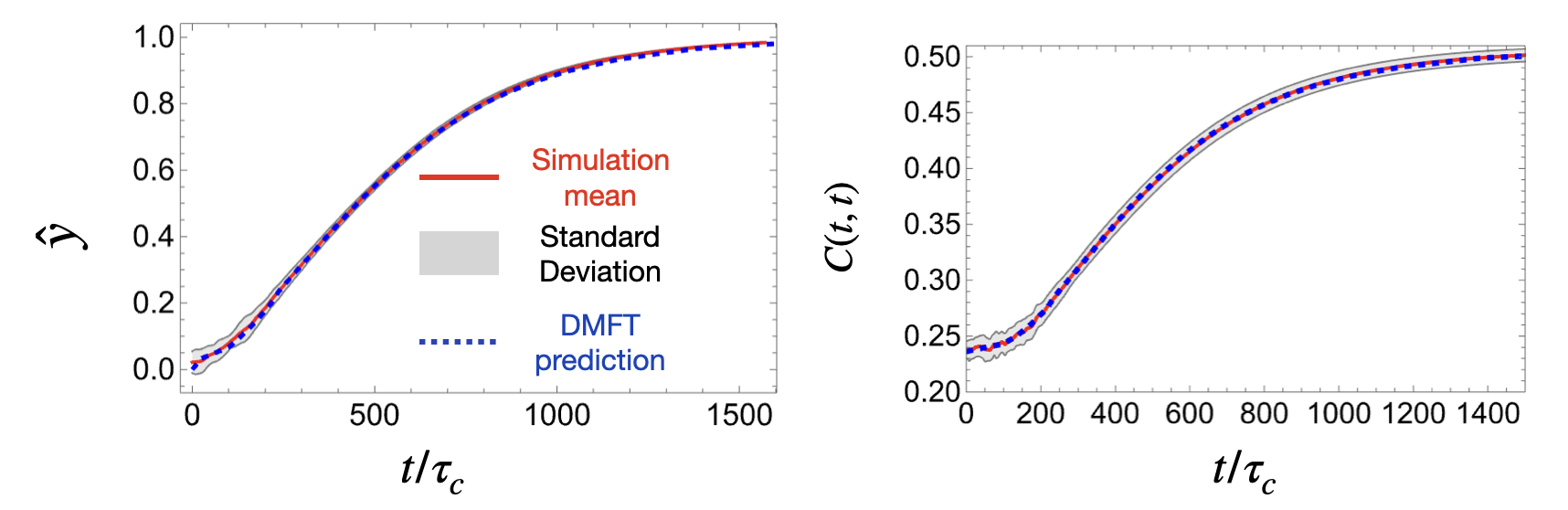}
\caption{Mean and standard deviation over 33 realizations for the network output (left) and diagonal correlation C(t,t) (right), after alignment by the realization-dependent onset time. The dashed curves show the DMFT predictions.}
\label{fig:mean}
\end{figure}

We now shift each run by its corresponding onset time. We determine the onset times of the various runs by demanding that they reach the output $\hat y =0.8$ at the same time. In Fig.\ref{fig:shifted}, we show the overlap of the 5 representative runs shifted by their onset times, along with the DMFT prediction. We also show the critical time $t_{\text{cr}}$ and the correspoding critical value of the feedback $\hat y_c$ as predicted by DMFT, which does agree quite well with the data. Finally we plot the mean and the standard deviation for 33 runs in Fig.\ref{fig:mean} for the network output and the diagonal correlation matrix $C(t,t)$. We find excellent agreement with the DMFT prediction which validates the theory.
\begin{figure}
\centering
\includegraphics[width=0.85\linewidth]{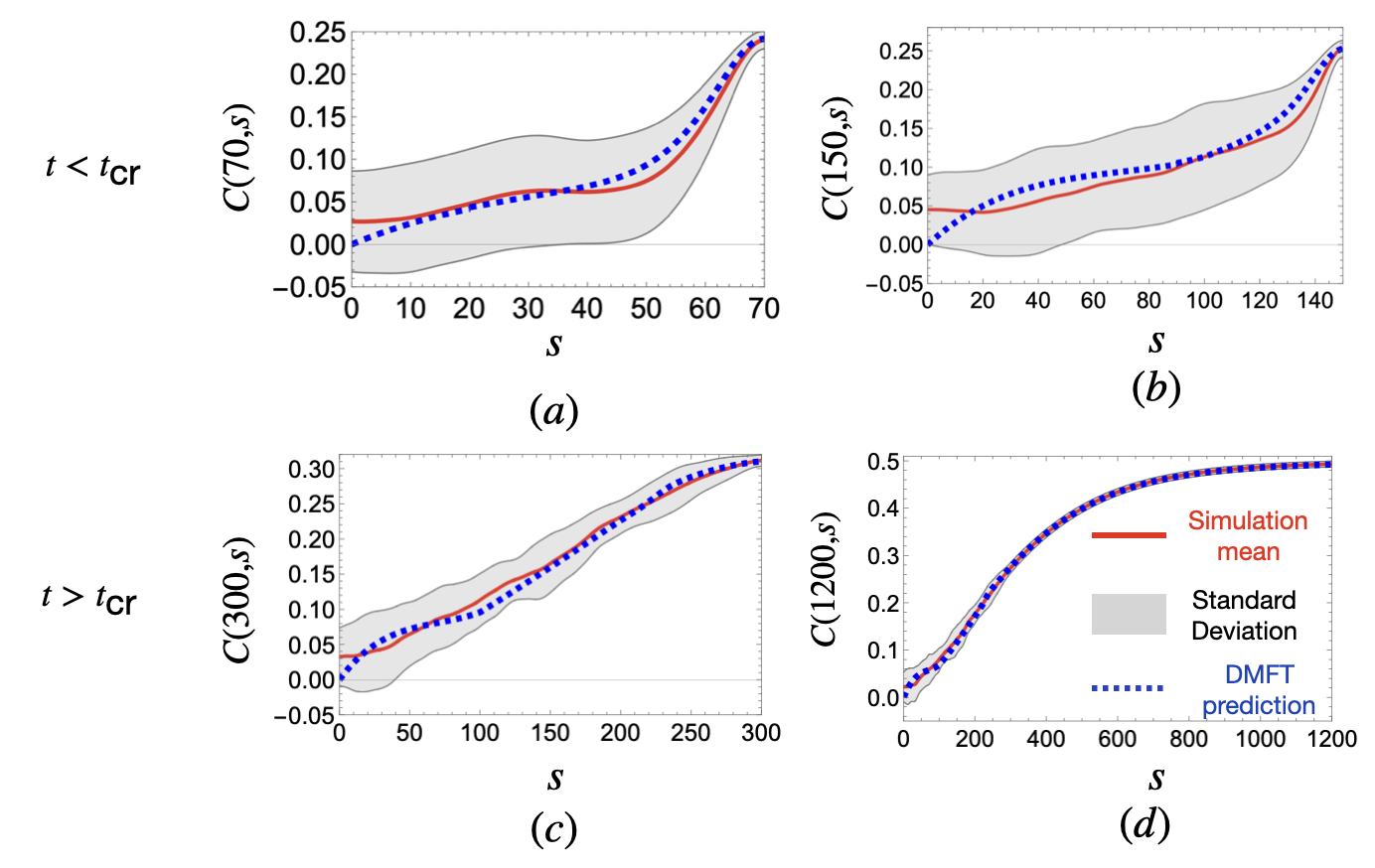}
\caption{Off-diagonal elements of the correlation matrix $C(t,s)$ along with the standard deviation for representative times before ((a) and (b)) and after ((c) and (d)) critical time. Dashed curves are the DMFT predictions.}
\label{fig:OffD}
\end{figure}
Finally, we plot the off diagonal correlation function $C(t,s)$ for representative choices of t and over $t-s \in\{0, t\}$ in Fig.\ref{fig:OffD}. We look at two slices of the off-diagonal elements for $t< t_{\text{cr}}$ (Fig.\ref{fig:OffD}(a) and (b)). As predicted by the DMFT,   in this chaotic regime, as we move away from the diagonal, we clearly see a fast decay to a plateau over a time scale $t_{\text{decay}} \sim 20 \tau_c$ followed by slow evolution of the plateau to 0 at s=0. The off-diagonal elements are noisier especially in the chaotic regime, but the mean agrees quite well with our approximation.  
For $t> t_{\text{cr}}$, the fast decay is absent, signifying frozen fast dynamics and we only see a slow learning induced evolution of the plateau. 
\section{Conclusion}
\label{sec:Conclusion}
In this work, we have developed a non-equilibrium dynamical mean-field theory for recurrent neural networks undergoing slow, feedback-driven learning. Going beyond  characterizing the final trained network, we followed the evolution of the network dynamics throughout learning. This allowed us to describe how plasticity progressively reorganizes the statistical state of the recurrent network and drives a transition between dynamical regimes.
A central result of our analysis is that learning generates an evolving effective feedback strength that acts as a dynamical control parameter. As the learned feedback develops, the effective dynamical landscape is progressively deformed, driving the network from an initially chaotic regime towards a dynamically stable state. The theory identifies a well-defined critical feedback strength and a corresponding learning time marking the transition between these regimes. By retaining the full two-time structure of the correlation function, rather than assuming stationarity, the theory captures both the rapid relaxation of fluctuations and the slow evolution of the correlation plateau associated with the learning process.

The transition identified here is a non-equilibrium dynamical transition rather than an equilibrium thermodynamic transition. The effective potential provides a convenient representation of the dynamical mean-field equations and should not be interpreted as a thermodynamic free energy. The two dynamical phases are distinguished by the presence or absence of persistent fast fluctuations and are separated by a bifurcation of the self-consistent DMFT solutions.

We further demonstrated that the resulting DMFT predictions quantitatively describe the learning dynamics observed in finite-size simulations. In particular, after accounting for realization-dependent fluctuations in the onset time, the ensemble-averaged trajectories and two-time correlations agree with the theoretical prediction. These finite-size variations are consistent with fluctuations around the macroscopic learning trajectory described by the DMFT, but instead lead to fluctuations in the time at which individual realizations commence macroscopic learning.

More broadly, this work suggests that dynamical mean-field theory can provide a useful framework for connecting microscopic plasticity rules to macroscopic changes in the dynamical state of recurrent networks. The two-time description developed here makes it possible to characterize learning trajectories beyond the stationary states typically considered in analyses of trained networks, and provides a route toward studying how different forms of plasticity shape the temporal organization of recurrent computation.

An important direction for future work is to extend this framework to learning rules that modify the recurrent connectivity itself, rather than only the readout and feedback pathways considered here. Such plasticity would allow the network to develop persistent internal structure and long-term memory, raising the question of how previously acquired dynamical states influence subsequent learning dynamics and the ability of the network to acquire new tasks. Extending the present non-equilibrium DMFT framework to this setting could provide a way to study the interaction between memory formation, dynamical phase transitions, and continual learning in recurrent networks.

\section*{Acknowledgements}
V.V. would like to thank Dr. Rodrigue Rizk for his valuable comments on the manuscript. V.V. is supported by startup funds from the University of South Dakota and by the U.S. Department of Energy, EPSCoR program under contract No. DE-SC0025545.

\appendix

\section{Derivation of DMFT equation}
\label{app:DMFT}

The generating functional which enforces the equation of motion of the neurons can be written as a path integral over the neuron degrees of freedom:
\begin{equation}
Z(J, W^{fb}) = \int \prod_i \mathcal{D}x_i \mathcal{D}\tilde{x}_i \; \exp \Bigg\{
i \sum_i \int dt \, \tilde{x}_i(t) \Big[ \tau_c \dot{x}_i(t) + x_i(t) - \sum_j \left(J_{ij}+W_i^{fb}w_j^{out}\right) \phi(x_j(t))  \Big] 
\Bigg\}
\end{equation}

we can perform an ensemble average over the network parameters $J_{ij}, W_i^{fb}$, using the the distributions 
\begin{eqnarray}
    P(J_{ij}) = \frac{g}{\sqrt{2\pi N}}\exp{\{-\frac{N}{2g^2}J_{ij}^2\} }, \ \ \ \  P(W_i^{fb}) = \frac{\sigma_{\text{fb}}}{\sqrt{2\pi N}}\exp{\Big\{-\frac{N}{2\sigma_{\text{fb}}^2}(W^{fb}_{i})^2\Big\} }
\end{eqnarray}

\begin{eqnarray}
     \bar Z &= &  \int DJ \int D W^{fb} P(J) P(W^{fb})Z(J, W^{fb}) \nonumber\\
     &=& \int \prod_i \mathcal{D}x_i \mathcal{D}\tilde{x}_i \; \exp \Bigg\{
i \sum_i \int dt \, \tilde{x}_i(t) \Big[ \tau_c \dot{x}_i(t) + x_i(t)\Big] \nonumber\\
&- &\sum_i\int dt \int dt'\tilde x_i(t) \tilde x_i(t')\Big[ \sum_j \frac{g^2}{2N}\phi(x_j(t)) \phi(x_j(t')) +\frac{\sigma^2_{fb}}{2N}y(t)y(t')  \Big] \Bigg\}
\end{eqnarray}
The path integral is done using a saddle point approximation, where the saddle point is given by 
\begin{eqnarray}
    \tilde x_i^*(t) &= & i \int ds \bar C^{-1}(t,s) (\dot x(s) \tau_c+x(s)) \ \  \text{where}  \ \ \bar C^{-1}(t,s) \ \ \text{solves} \nonumber\\
    && \int ds \bar C(t,s) \bar C^{-1}(s,u) = \delta(t-u) \ \ \ \text{and} \nonumber\\ 
     \bar C(t,s) &=&  \frac{g^2}{N} \sum_j \phi(x_j(t)) \phi(x_j(s))+ \frac{\sigma^2_{\text{fb}}}{N}y(t) y(s) 
\end{eqnarray}
The saddle point is on the imaginary axis but the integral over $\tilde x_i(t)$ is along the real line. We therefore choose a contour that passes through the saddle point and orients along the path of steepest descent which in this case points along  direction of the real line. So we simply shift the contour to be parallel to the real line passing through $\tilde x_i(t)^*$ noting that we do not cross any poles and the contributions from the segments at infinity vanish. This yields the result 
\begin{eqnarray}
    \bar Z = \prod_{i} \int Dx_i \exp \Bigg\{
i \sum_i \int dt  ds \Big[ \tau_c \dot{x}_i(t) + x_i(t)\Big]\bar C^{-1}(t,s)\Big[ \tau_c \dot{x}_i(s) + x_i(s)\Big]\Bigg\}
\end{eqnarray}
Next we introduce conjugate field $\hat C (t,s)$ which enforces the definition for the collective field $\bar C(t,s)$,
\begin{eqnarray}
    \bar Z &=& \int D \hat C D \bar C \exp \Bigg\{
i \sum \int dt  ds \bar C(t,s) \hat C(t,s) \Bigg\} \nonumber\\
&& \prod_i  D \eta_i \exp \Bigg\{-\frac{1}{2}\sum_i\int dt ds \eta_i(t) \bar C^{-1}(t,s) \eta_i(s) -i \int dt ds \hat C(t,s)\Big[\frac{g^2}{N}\sum_j \phi(x_j(t))\phi(x_j(s)) + \frac{\sigma^2_{\text{fb}}}{N}y(t)y(s)\Big]\Bigg\} \nonumber \\
& \equiv & \int D \hat C D \bar C \exp \Bigg\{
i \sum \int dt  ds \bar C(t,s) \hat C(t,s) \Bigg\} \exp \Bigg\{ \ln \hat Z(\bar C, \hat C) \Bigg\}
\end{eqnarray}
Next we find the saddle point solution by minimizing the action over $\hat C$ which yields 
\begin{eqnarray}
    \bar C(t,s) = \frac{g^2}{N}\sum_i\langle \phi(x_i(t))\phi(x_i(s)) \rangle_{\hat Z(\bar C , \hat C)} +  \frac{\sigma^2_{\text{fb}}}{N}\langle y(t)y(s) \rangle_{\hat Z(\bar C , \hat C)}
\label{eq:Cavg}
\end{eqnarray}
i.e., the matrix $\bar C$ is set to its average value computed self consistently. When we substitute this back into our path integral, using the central limit theorem, we can ignore the variance in $\bar C$ in the large N limit, which then effectively reduces our generating functional upto a normalizing factor to 
\begin{eqnarray}
\hat Z= \prod_i  D \eta_i \exp \Bigg\{-\frac{1}{2}\sum_i\int dt ds \eta_i(t) \bar C^{-1}(t,s) \eta_i(s) \Bigg\} 
\end{eqnarray}
where $\bar C (t,s)$ is average value in Eq.~\ref{eq:Cavg} that is determined self consistently using the Gaussian action $\hat Z$. 

\bibliographystyle{utphys.bst}
 \bibliography{RNN}

\providecommand{\href}[2]{#2}\begingroup\raggedright\begin{thebibliography}{10}

\bibitem{sompolinsky1988chaos}
H.~Sompolinsky, A.~Crisanti, and H.~Sommers, ``Chaos in random neural
  networks,'' {\em Physical Review Letters} {\bfseries 61} no.~3, (1988)
  259--262.

\bibitem{Dauce1998}
E.~Dauc{\'e}, M.~Quoy, B.~Cessac, B.~Doyon, and M.~Samuelides,
  ``Self-organization and dynamics reduction in recurrent networks: stimulus
  presentation and learning,''
  \href{http://dx.doi.org/10.1016/S0893-6080(97)00131-7}{{\em Neural Networks}
  {\bfseries 11} no.~3, (1998) 521--533}.

\bibitem{LajeBuonomano2013}
R.~Laje and D.~V. Buonomano, ``Robust timing and motor patterns by taming chaos
  in recurrent neural networks,'' \href{http://dx.doi.org/10.1038/nn.3405}{{\em
  Nature Neuroscience} {\bfseries 16} (2013) 925--933}.

\bibitem{SussilloAbbott2009}
D.~Sussillo and L.~F. Abbott, ``Generating coherent patterns of activity from
  chaotic neural networks,''
  \href{http://dx.doi.org/10.1016/j.neuron.2009.07.018}{{\em Neuron} {\bfseries
  63} no.~4, (2009) 544--557}.

\bibitem{MastrogiuseppeOstojic2018}
F.~Mastrogiuseppe and S.~Ostojic, ``Linking connectivity, dynamics, and
  computations in low-rank recurrent neural networks,''
  \href{http://dx.doi.org/10.1016/j.neuron.2018.07.003}{{\em Neuron} {\bfseries
  99} no.~3, (2018) 609--623.e29}.

\bibitem{RivkindBarak2017}
A.~Rivkind and O.~Barak, ``Local dynamics in trained recurrent neural
  networks,'' \href{http://dx.doi.org/10.1103/PhysRevLett.118.258101}{{\em
  Physical Review Letters} {\bfseries 118} no.~25, (2017) 258101}.

\bibitem{Clark2026}
D.~G. Clark, B.~Bordelon, J.~A. Zavatone-Veth, and C.~Pehlevan, ``Structure,
  disorder, and dynamics in task-trained recurrent neural circuits,''
  \href{http://dx.doi.org/10.64898/2026.03.02.708943}{{\em bioRxiv} (2026) }.

\bibitem{Jaeger2001}
H.~Jaeger, ``The ``echo state'' approach to analysing and training recurrent
  neural networks,'' Tech. Rep. 148, German National Research Center for
  Information Technology, 2001.

\bibitem{Koryakin2012}
D.~Koryakin, J.~Lohmann, and M.~V. Butz, ``Balanced echo state networks,''
  \href{http://dx.doi.org/10.1016/j.neunet.2012.08.008}{{\em Neural Networks}
  {\bfseries 36} (2012) 35--45}.

\bibitem{Murray2019}
J.~M. Murray, ``Local online learning in recurrent networks with random
  feedback,'' \href{http://dx.doi.org/10.7554/eLife.43299}{{\em eLife}
  {\bfseries 8} (2019) e43299}.

\bibitem{Miconi2017}
T.~Miconi, ``Biologically plausible learning in recurrent neural networks
  reproduces neural dynamics observed during cognitive tasks,''
  \href{http://dx.doi.org/10.7554/eLife.20899}{{\em eLife} {\bfseries 6} (2017)
  e20899}.

\bibitem{AsabukiClopath2025}
T.~Asabuki and C.~Clopath, ``Taming the chaos gently: a predictive alignment
  learning rule in recurrent neural networks,''
  \href{http://dx.doi.org/10.1038/s41467-025-61309-9}{{\em Nature
  Communications} {\bfseries 16} (2025) 6784}.

\bibitem{MartinSiggiaRose1973}
P.~C. Martin, E.~D. Siggia, and H.~A. Rose, ``Statistical dynamics of classical
  systems,'' \href{http://dx.doi.org/10.1103/PhysRevA.8.423}{{\em Physical
  Review A} {\bfseries 8} (1973) 423--437}.

\bibitem{Janssen1976}
H.~K. Janssen, ``On a lagrangean for classical field dynamics and
  renormalization group calculations of dynamical critical properties,''
  \href{http://dx.doi.org/10.1007/BF01316547}{{\em Zeitschrift f{\"u}r Physik
  B} {\bfseries 23} (1976) 377--380}.

\bibitem{DeDominicis1976}
C.~De~Dominicis, ``Technics of field renormalization and dynamics of critical
  phenomena,'' {\em Journal de Physique Colloques} {\bfseries 37} (1976)
  C1--247--C1--253.

\bibitem{CrisantiSompolinsky2018}
A.~Crisanti and H.~Sompolinsky, ``Path integral approach to random neural
  networks,'' \href{http://dx.doi.org/10.1103/PhysRevE.98.062120}{{\em Physical
  Review E} {\bfseries 98} (2018) 062120}.

\end{thebibliography}\endgroup
\end{document}